\documentclass[conference]{IEEEtran}
\usepackage{amsmath,amssymb,amsfonts}
\usepackage{graphicx}
\usepackage{booktabs}
\usepackage{multirow}
\usepackage{color}
\usepackage{caption}
\usepackage{enumitem}
\usepackage{siunitx}
\usepackage{float}    
\usepackage{adjustbox}   
\usepackage[table]{xcolor} 
\usepackage{xurl}
\usepackage[colorlinks=true, urlcolor=black, linkcolor=black, citecolor=black]{hyperref}
\usepackage[capitalize]{cleveref}

\usepackage{geometry}
\usepackage{fancyhdr} 

\IEEEoverridecommandlockouts

\title{Context-Aware Zero-Shot Anomaly Detection in Surveillance Using Contrastive and Predictive Spatiotemporal Modeling}

\author{
\IEEEauthorblockN{\small Md. Rashid Shahriar Khan\IEEEauthorrefmark{1}, Md. Abrar Hasan\IEEEauthorrefmark{2}, Mohammod Tareq Aziz Justice\IEEEauthorrefmark{3}}
\IEEEauthorblockA{\small Department of Computer Science and Engineering, BRAC University, Dhaka, Bangladesh\\
Email:
\IEEEauthorrefmark{1}rashid.shahriar.khan@g.bracu.ac.bd, 
\IEEEauthorrefmark{2}md.abrar.hasan@g.bracu.ac.bd, 
\IEEEauthorrefmark{3}mohammod.tareq.aziz@g.bracu.ac.bd}
}

\begin{document}
\maketitle
\thispagestyle{fancy}
\section{Abstract}
Detecting anomalies in surveillance footage is inherently challenging due to their unpredictable and context-dependent nature. This work introduces a novel context-aware zero-shot anomaly detection framework that identifies abnormal events without exposure to anomaly examples during training. The proposed hybrid architecture combines TimeSformer, DPC, and CLIP to model spatio-temporal dynamics and semantic context. TimeSformer serves as the vision backbone to extract rich spatial-temporal features, while DPC forecasts future representations to identify temporal deviations. Furthermore, a CLIP-based semantic stream enables the detection of concept-level anomalies through context-specific text prompts. These components are jointly trained using InfoNCE and CPC losses, aligning visual inputs with their temporal and semantic representations. A context-gating mechanism further enhances decision-making by modulating predictions with scene-aware cues or global video features. By integrating predictive modeling with vision language understanding, the system can generalize to previously unseen behaviors in complex environments. This framework bridges the gap between temporal reasoning and semantic context in zero-shot anomaly detection for surveillance. The code for this research has been made available at this link \href{https://github.com/NK-II/Context-Aware-Zero-Shot-Anomaly-Detection-in-Surveillance}{\textit{\small https://github.com/NK-II/Context-Aware-Zero-Shot-Anomaly-Detection-in-Surveillance}}.    
\section{Introduction}
\label{sec:intro}

Anomaly detection is a fundamental aspect of various domains, including finance, healthcare, cybersecurity, etc., where identifying irregular patterns is crucial for ensuring security and operational efficiency. In the realm of surveillance in crowd areas, anomaly detection is particularly important for identifying potential threats. Since surveillance activity is rapidly growing in the world day by day, it has become a crucial demand to analyze videos through automated systems. As a result, video anomaly detection (VAD) has emerged as an important research area that leverages computational models to automate the identification of abnormal events in security videos. Ensuring public safety in environments such as streets, campuses, airports, and shopping malls heavily relies on effective VAD systems that can alert authorities to crimes, accidents, or other irregular behaviors as they occur. In the early stage, when the video anomaly detection system was newly introduced to us, it used to focus on manually crafted features and statistical models. Traditional computer vision algorithms used motion trajectories, object speed, or pixel-level changes to characterize normal patterns, raising alerts when deviations occur. However, these old systems had limitations as they were hand-crafted. For example, capturing complex spatiotemporal patterns in crowded or dynamic scenes was a major challenge where these approaches were not suitable. After that, a revolutionized anomaly detection technique called deep learning emerged, with the capability of learning from rich feature representations directly from data. These deep learning models automatically learn what defines normal behavior in a video, making them suitable for detecting subtle or complex anomalies that traditional algorithms often miss.

Despite advances, video anomaly detection systems still face significant challenges in efficiency and real-world applicability. A key limitation is the lack of context awareness and zero-shot generalization. Many existing methods treat all environments uniformly, ignoring that normal behavior varies with context like time or events. This leads to missed detections or false alarms in dynamic, long-term surveillance—for example, a crowd might be normal at one time but anomalous at another. Additionally, most deep learning models struggle to detect novel anomalies unseen during training, as they rely on patterns implicit in the normal data. This causes poor performance with subtle or context-dependent anomalies, such as weather-related changes. Given the rarity and diversity of abnormal events, predefining all anomalies in training is infeasible, making anomaly detection essentially a zero-shot learning problem where models must identify unknown anomalies without prior examples.

The primary objective of this research is to design a novel deep learning model for surveillance videos that incorporates context-conditioning. The model should learn representations of normal behavior not in isolation, but in relation to contextual factors. This involves creating a dual-stream architecture where one stream models the video’s spatiotemporal content and another stream provides contextual input, enabling the system to adjust its understanding of normality based on context. Secondly, leveraging contrastive learning and external semantic knowledge to allow the detection of anomalies that were never explicitly observed during training. Specifically, integrate a vision-language model like CLIP to imbue the system with semantic understanding, so that if an anomalous event corresponds to a known concept, the model can recognize it via textual descriptions even without direct training examples. 

Additionally, use self-supervised predictive modeling to learn generalizable temporal patterns that can flag novel deviations. Thirdly, utilizing a TimeSformer (Time-based Transformer) to capture spatial and temporal features of video sequences in a unified framework. Also, employing a contrastive predictive coding loss to train the model to predict future frame representations, reinforcing the learning of normal spatiotemporal dynamics. By combining contrastive learning (for context and semantic alignment) with predictive modeling (for temporal consistency), to ensure that the learned representation is robust to variations and sensitive to meaningful deviations. Finally, evaluating the developed model on both public benchmark datasets and a context-rich surveillance dataset rigorously. 

Key performance indicators include frame-level and video-level anomaly detection accuracy (e.g., measured by area under the ROC curve), as well as the false alarm rate in various contexts. So, our objective is to compare the performance against state-of-the-art anomaly detection methods (both context-agnostic and context-aware variants) to demonstrate improvements. Performing statistical analyses to verify the significance of performance gains and to quantify the contributions of each component (context conditioning, contrastive loss, predictive coding, etc.) to the overall system. Achieving these objectives will result in a context-aware, zero-shot anomaly detection system that advances the field of intelligent surveillance and has practical implications for deploying artificial intelligence in real-world security environments.

\section{Related Works}
\label{sec:related works}

Early approaches to surveillance anomaly detection were often based on \textbf{manually crafted features} and assumptions about typical motion patterns. For example, techniques like \textbf{trajectory analysis} and \textbf{optical flow} were widely used to model normal movement within a scene, with any deviations from these patterns considered anomalies \cite{Luo2017A}. Traditional methods often relied on statistical models that analyzed features such as object speed, direction, and distances between objects. 

Addressing practical implementation, Sharma et al. \cite{Sharma2021A} proposed a broader solution featuring not only a CNN-LSTM model but also a well-defined \textbf{system architecture}. This approach yielded the highest accuracy of 98.87\% and innovatively introduced a \textbf{mobile application} using scalable cloud services to notify responsible personnel. A significant drawback, however, was the dataset, which consisted only of data from movies, potentially limiting its future real-world performance \cite{Sharma2021A}. Other works focused on improving model effectiveness in complex scenarios; for instance, Nasaruddin et al. \cite{Nasaruddin2020Deep} introduced a hybrid of \textbf{Background Subtraction} with a \textbf{3D CNN} model, incorporating Visual Attention algorithms to enhance efficiency in diverse environments, addressing issues where previous models failed in highly crowded places. 

Another line of research explored \textbf{generative models} for anomaly detection. Mahdyar Ravanbakhsh et al. \cite{Ravanbakhsh2017Abnormal} developed a model using a double conditional \textbf{GAN} trained exclusively on normal data. This model identifies anomalies by its inability to reconstruct unseen, abnormal regions in data frames. The approach, however, struggles to capture small or obscure abnormal objects. Concurrently, the limitations of CNNs, specifically their limited receptive field, led researchers to explore \textbf{Transformer-based architectures}. Transformers can extract features across different time steps on a wide range of inputs \cite{Qaraqe2024Crowd}\cite{Aslam2024TransGANomaly}\cite{Sultani2018Real}. 

While a Transformer-based GAN achieved satisfactory results, it was not suitable for \textbf{real-time application} and faced challenges with computational overhead and occluded objects \cite{Yang2022A}\cite{Radford2021Learning}. The high computational costs and limitations of earlier models spurred the development of \textbf{Zero-Shot Learning (ZSL)} techniques. The \textbf{ALFA} framework introduced by Zhu et al. \cite{Zhu2024Do} utilizes \textbf{Run-time Prompts (RTP)} to generate context-aware prompts, achieving a 93.2\% AUROC. Another method synthesized pseudo-outliers from inlier data, eliminating the need for real anomaly samples during training and improving AUC by 15-20\% \cite{Jeong2023WinCLIP}. Vision-language models like \textbf{CLIP} became pivotal. The \textbf{WinCLIP} model, proposed by Jeongheon et al. \cite{Jeong2023WinCLIP}, leverages CLIP for scalable zero-shot and few-shot anomaly classification, achieving up to 95.2\% image-level AUROC. Other models like \textbf{PromptAD} \cite{Li2024PromptAD} and \textbf{AnomalyCLIP} \cite{Zhou2023AnomalyCLIP} also integrated text prompts with CLIP to detect anomalies without requiring target-domain training data. Further research has extended into \textbf{Generalized Zero-Shot Learning (GZSL)}, which tests on both seen and unseen classes, with one study achieving a harmonic mean of 65.3\% by using a \textbf{CBAM} attention mechanism \cite{Liu2024A}. 

The most recent developments focus on \textbf{context-awareness}. Sun and Gong (2023) \cite{Sun2023Hierarchical} proposed a scene-aware technique using hierarchical semantic contrast, and Yang and Radke (2024) \cite{Yang2024Context} introduced the \textbf{Trinity} framework to model appearance, motion, and context. This thesis aims to synthesize these threads, focusing on the success of \textbf{contrastive learning} and \textbf{predictive modeling} to create a contextually informed, zero-shot anomaly detection model suitable for real-world requirements.

\section{Methodology}
Our proposed method offers a zero-shot approach to anomaly detection in surveillance videos, requiring no abnormal samples during training. The model is designed to be both context-aware and temporally predictive by leveraging a dual-stream architecture. One stream uses a transformer-based model, TimeSformer \cite{Qaraqe2024Crowd}\cite{Yang2022A}\cite{Bertasius2021Is}, to extract spatiotemporal features from video input. The other processes contextual metadata—such as time, day, or scene descriptions—using a text encoder like CLIP, generating semantic embeddings. These two streams are fused to create a joint representation that reflects the scene's behavior within its specific context. 

The training employs two complementary loss functions: a contrastive loss that aligns matching video-context pairs while separating mismatches, and a predictive loss based on Contrastive Predictive Coding (CPC) \cite{Liu2024A}, which encourages the model to forecast future video features. At inference, the system produces a context alignment score and a predictive score; mismatches in either signal potential anomalies. This dual-scoring mechanism enables the detection of both contextual anomalies (e.g., people in restricted areas at odd hours) and temporal anomalies (e.g., sudden or unusual movements). Using CLIP’s vision-language capabilities further enhances generalization, allowing detection of novel behaviors not seen during training.

Our system is trained entirely on normal video segments paired with their proper context where it never sees any abnormal data while training. This is what makes it zero-shot, it learns only what is normal and identifies anything different as abnormal. During application, the model generates two scores. The first is a context alignment score that measures how well the video matches the expected context. The second is a predictive score that reflects how well the model could anticipate the next part of the video. 

A low context alignment score or a high prediction error signals an anomaly. These scores can be combined into a final anomaly score to make the decision. This combination of techniques enables the model to identify both context-based anomalies, like people appearing in places they shouldn’t be at that time, and temporal anomalies, like unusual actions or motion patterns. The use of a pre-trained vision language model like CLIP further helps the model recognize concepts it has never seen during training by linking visual patterns to high-level ideas. As a result, the system is able to generalize well and flag unexpected events without needing labeled examples of anomalies. This method provides a robust and flexible solution for real-world anomaly detection in surveillance settings. 


\subsection{Proposed Model Architecture}
The proposed Context-Aware Zero-Shot Anomaly Detection framework is a unified model that jointly learns spatiotemporal representations and semantic context for surveillance videos. The architecture consists of four main components, a global scene encoder for capturing holistic video context, a predictive modeling module for anticipating future scene dynamics, a context-conditioning network to modulate predictions based on scene context, and a CLIP-based text encoder to inject high-level semantic knowledge. These components are trained together in an end-to-end manner to learn a shared embedding space for video and textual context, enabling zero-shot inference. 

Figure~\ref{fig:VideoPipeline} illustrates the architecture. Input video frames are processed by the TimeSformer-based encoder, whose outputs feed into the DPC-RNN predictive module. A context-conditioning subnetwork takes global scene features or associated text descriptions to produce modulation parameters that inform the predictive module. In parallel, a text encoding branch using CLIP’s language encoder provides a semantic context vector. All feature streams are projected into a common embedding space and optimized with a contrastive InfoNCE loss and a hybrid Contrastive Predictive Coding (CPC) loss \cite{VanDenOord2018Representation}. This design allows the model to learn what constitutes “normal” patterns in a scene and detect deviations as anomalies without explicit anomaly examples, in a zero-shot way. Next, we detail each module and the training methodology. \\

\subsubsection{Zero-Shot Learning (ZSL)}
What if a machine learning model could learn and work without ever needing labeled data? This is precisely where Zero-Shot Learning comes into play. It is a machine learning algorithm that enables a model to identify or classify new instances of data or concepts that it had not previously seen or was not explicitly trained on such labeled data as well as having the ability handle data for which it wasn't particularly trained. It allows the model to recognize things that it had never encountered. It utilizes the information it already knows and connects it to the new situation, which was presented in front of it. 

Let me explain this with an example: Suppose a model is trained to recognize animals but not particular ones like Zebra in its training phase. In case of other machine learning approaches, the model will fail to classify an animal being Zebra because it was never explicitly shown any labeled data of Zebra during its training period. But ZSL will be able to figure it out by using description even though it was also not shown any examples of it. The ZSL model knows about the information of animals from its training phase like 'they have four legs,' 'has stripes' or 'has a tail'. The ZSL model is given a new description that reads as, 'A horse is an animal with black and white stripes and a tail that lives in Africa.' Using its knowledge on animal attributes and characteristics along with the given description it will successfully deduce that the animal is in fact a Zebra even though it had no prior knowledge about it during its training. This is the power of Zero-Shot Learning. It uses these auxiliary information to its advantage. 

\subsubsection{Contrastive Predictive Coding (CPC)}
This is an unsupervised learning approach that extracts useful representations from high-dimensional data. The key idea behind this is to learn such representation by predicting the future latent feature representation through the use of autoregressive models. The key insight here is that using past contexts, CPC predicts future parts of a sequence but not by directly generating the data but by distinguishing the true future from the false ones, a learning procedure known as contrastive learning. In this case, a probabilistic contrastive loss is used alongside negative sampling. It learns rich temporal data without labels, which is ideal for temporal modeling. \textbf{Equation~\ref{eq:Combined Loss}}, \textbf{$\mathcal{L}_{align + pred} / \mathcal{L}_{total}$}, shows how we combined these two losses for our particular task.

\begin{figure}[H]
\centering
\includegraphics[width=\linewidth]{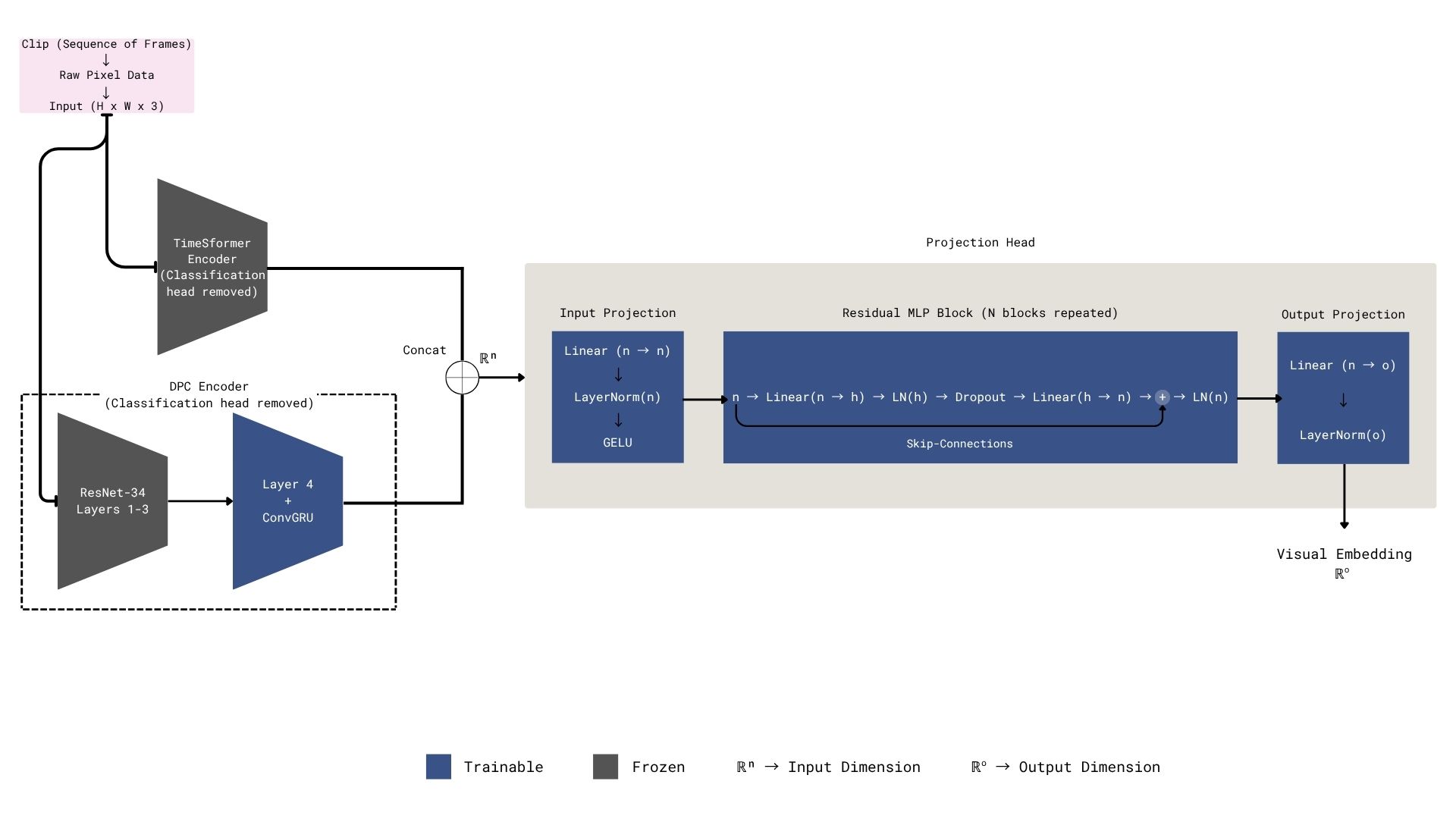}
\caption{Video Architecture Pipeline}
\label{fig:VideoPipeline}
\end{figure}

\subsubsection{TimeSformer for Spatiotemporal Features}
To capture the overall appearance and movement patterns in surveillance videos, we use TimeSformer, a transformer-based model designed specifically for video understanding\cite{Aslam2024TransGANomaly}. TimeSformer treats a video not as a single image sequence but as a series of smaller visual patches. Each frame in the video is split into patches, which are then turned into vectors through linear projection\cite{Deshpande2023Anomaly}. To help the model understand the position and order of these patches, we add spatial and temporal position information to each one.\\

These processed patches form a long sequence of tokens that are passed through multiple transformer layers. Through self-attention mechanisms, the model learns how different patches relate to each other across both space and time. This helps the model capture long-range dependencies and understand the broader context of the video clip. We use a specific variant of TimeSformer known as “divided space-time attention,” which separates the spatial and temporal attention computations to make the model more efficient, especially for longer video clips. The result of this processing is a rich sequence of feature vectors that describe the video at each moment. In our implementation, we use these outputs in two main ways. First, the feature from each frame is passed to a predictive model that learns how the scene evolves over time. Second, these features are mapped into a shared embedding space so that they can be compared directly with semantic context information from text descriptions.\\

By using TimeSformer, our model can pay attention to all parts of a scene at once. This is especially important in anomaly detection, where unusual events often involve complex interactions or context-sensitive behavior. Unlike traditional convolutional networks which focus on local patterns, TimeSformer’s global self-attention enables it to understand the full picture. This makes it ideal for modeling what is “normal” in different surveillance scenarios.\\

\subsubsection{Temporal Predictive Modeling with DPC-RNN}
While the TimeSformer encoder extracts features from each frame in the video, it does not capture how the scene evolves over time. To model this temporal progression, we used a Dense Predictive Coding Recurrent Neural Network (DPC-RNN). This module takes the sequence of frame-level features produced by the TimeSformer and learns how to predict what should come next in a normal sequence of events. At every moment in time, the RNN updates its internal memory using the current feature and the previous hidden state. This updated state acts as a summary of everything the model has seen so far. The predictive head then uses this state to forecast future representations of the video. Instead of trying to generate actual video frames the model predicts the high-level features that represent the expected future. The goal is to match these predicted features as closely as possible with the actual ones that occur later in the sequence.\\

\textbf{Figure~\ref{fig:VideoPipeline}} illustrates how the video pipeline is structured in its entirety. Layer4 \& ConvGRU of DPC-RNN \cite{Hasan2016Learning} is fine-tuned for domain adaption. Two separate inputs enter both the feature extractor backbones, varying in terms of the number of frames that the backbones anticipate. Their initial output embeddings are concatenated and is then passed to the residual MLP projection block, where it will be projected to the CLIP's space. We used multiple residual blocks in the projection head, which will help make the training stable and the transmission of information and gradients easier. Furthermore, the Projection Head will learn to project its output into CLIP's space in such a way that visual embeddings will land closer to its corresponding textual embedding and further away from negatives. Firstly, inside the input projection, the concatenated visual embedding passes through a Linear layer, which is a FCNN or a Dense layer, and the layer will learn a Linear transformation of the input features. The output from this is then normalized using Layer Normalization, which helps to stabilize the training process. Also ensures that the network's layer receives data information with a consistent distribution with a mean of 0 and standard deviation of 1. A Gaussian Error Linear Unit (GELU) is used to introduce non-linearity into the model. The Residual MLP Block is the core processing unit of the Projection Head. As mentioned above, $N$ residual blocks are used, which helps the model to learn progressively more abstract and refined features of the input as a result of stacking of these blocks. The arrow from the input to the final output of this block that bypasses the inner layers, LN, activations is a `skip' or `residual' connection. This is the essence of a Residual Network (ResNet) which solves the vanishing gradient problem. If we trace the path inside this block, the output from the Input Projection block is passed through a Linear layer which projects it to a higher dimension $h$ allowing the model to learn more richer representation. This a $h$-dimensional hidden vector. LN is applied to it. Also a regularization method is used here, allowing the model to generalize better to unseen, new data. After that the $h$-dimensional hidden vector is projected back to the original dimension. A skip-connection allows for the original input to be added to the output of this final linear layer. The entire sequence of this block is repeated $N$ times. It is then normalized and passed to the Output Projection block as input. The input is passed to through another linear layer where its projected to the specific dimension and normalized and thus produces the visual embedding, where $\mathbf{v} \in \mathbb{R}^{d}$ .\\

The training strategy follows a contrastive approach which is using the infoNCE loss. The model learns to make its predicted future feature similar to the actual future feature while making it dissimilar from unrelated ones. These unrelated or negative examples are typically taken from other parts of the same video or from different videos in the same batch. This way, the model learns meaningful patterns that consistently appear in normal videos, while ignoring small random changes like noise or slight movement that don't indicate anomalies. By summing the predictive loss over multiple future steps and across time, the DPC-RNN becomes skilled at learning how scenes usually unfold \cite{Joo2022CLIP}. This is useful for detecting anomalies because during the training phase the model only sees normal video clips. If something unusual happens later it won’t be able to predict it accurately, resulting in a large difference between the predicted and actual features. This prediction error becomes a clear signal that something unexpected is happening in the video.\\

\subsubsection{Context-Conditioning Network for Scene-Aware Modulation}
A key innovation in our model is its ability to adjust its behavior based on the specific scene it is observing. In real world surveillance, what is considered “normal” varies greatly from one location to another \cite{Yang2024Context}. For example, heavy pedestrian activity might be typical in a shopping mall but highly unusual in a restricted area. To account for this, we use a Context-Conditioning Network that adjusts the model’s predictions according to scene-specific information.\\

\begin{figure}[H]
\centering
\includegraphics[width=\linewidth]{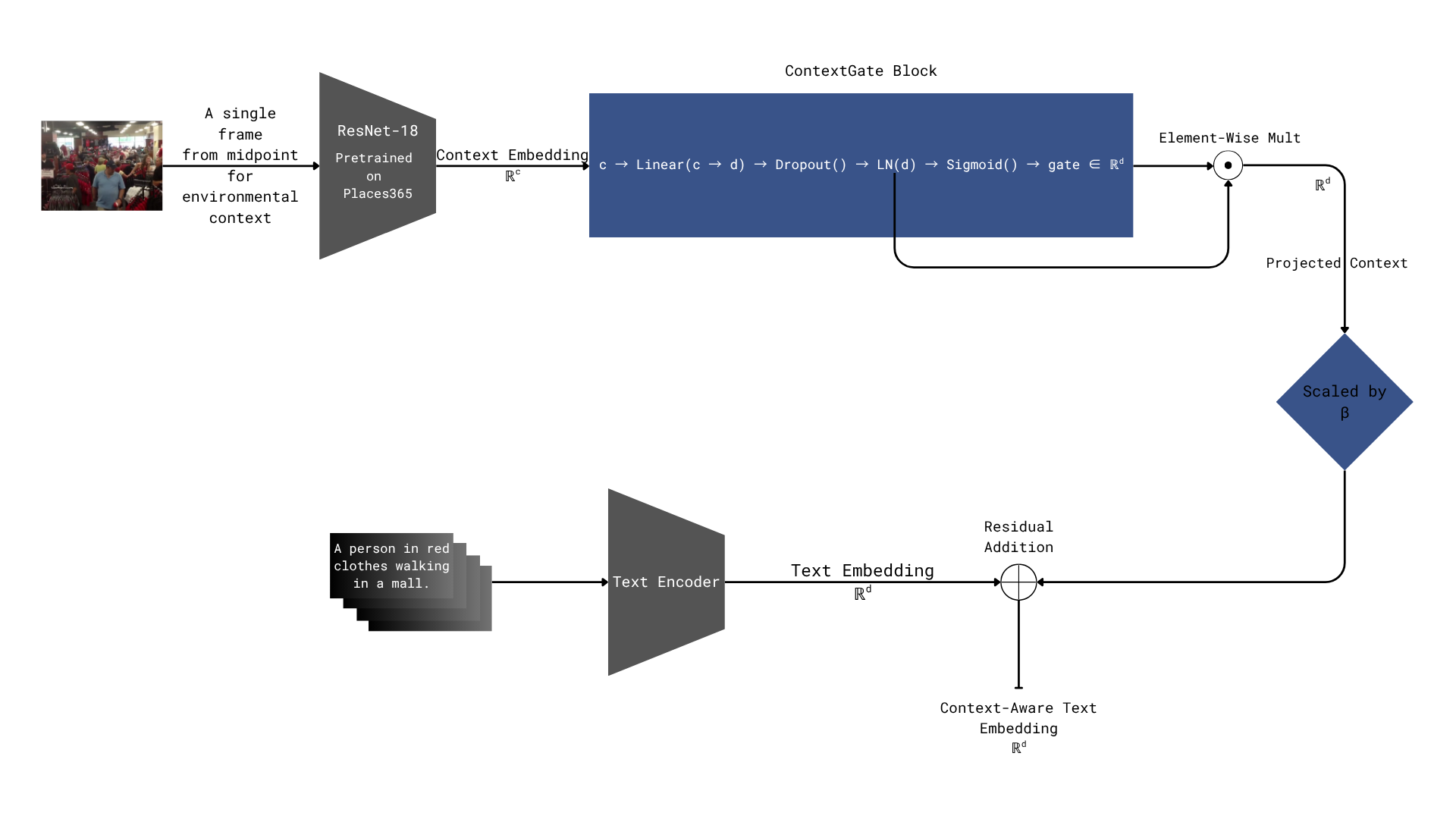}
\caption{Text Pipeline}
\label{fig:TextPipeline}
\end{figure}

The context used to guide the model can come from two sources, visual features and textual descriptions. Visually, the TimeSformer encoder captures static aspects of a scene, such as its layout or common background elements. Textually, we can provide a natural language description of the scene which is encoded using the CLIP text encoder. These inputs are processed to form a context vector that is used to adjust how the prediction model behaves. In our case, raw pixel data from a single mid-frame of a clip is provided to the Context-Conditioning Network as its input. It outputs a context vector embedding and is passed through a ContextGate Block which is FCNN or MLP. We introduced a learnable parameter here, \textbf{$\beta$}, as illustrate in \textbf{Figure~\ref{fig:TextPipeline}}, which is used to control how much information will be passed to the textual embedding so that not all its own information is washed out by the context vector. The ContextGate learns over time as to how much information is needed by the textual embedding to understand the surrounding context or whether it is capable of determining the context by itself or not. It is then fused with the text embedding through residual addition. Alternatively, the context vector can be directly added to the visual features before transferring it to the RNN. These adjustments help the model make scene-specific predictions and reduce false alarms.

This design enables the model to adapt to new environments without retraining, which is essential for zero-shot anomaly detection. It allows the system to understand what kind of behavior is expected in a given context and to detect deviations effectively.\\

\subsubsection{CLIP-Based Text Encoder for Semantic Context}
To enhance the model’s ability to understand high-level semantics and support zero-shot anomaly detection, we incorporate a CLIP-based text encoder. CLIP is a powerful model trained on a vast amount of image-text pairs. It can map both images and natural language descriptions into the same feature space, which allows for a meaningful comparison between visual and textual content.\\

In our system, we use the CLIP text encoder as shown in \textbf{Figure~\ref{fig:TextPipeline}} is used to convert descriptive sentences about a scene into feature vectors. These descriptions define what is typically expected in a given setting. Each sentence is transformed into a semantic embedding that represents the normal behavior or appearance of a specific scene. Rather than relying on CLIP’s image encoder, we train our TimeSformer-based video encoder to produce embeddings that align with these text representations. During training, we use contrastive learning to bring the video embedding closer to the matching context text and push it away from unrelated ones. This teaches the model to understand and align the video content with the semantic expectations.

This text-guided alignment gives our system strong zero-shot capabilities. Even if the model never saw a specific anomaly during training, it can still recognize it by detecting a mismatch between the video and the expected context description. The CLIP based text encoder acts as a form of semantic supervision that guides the model to understand what is normal and helps it to detect when something deviates in a meaningful way.

\begin{figure}[H]
\centering
\includegraphics[width=\linewidth]{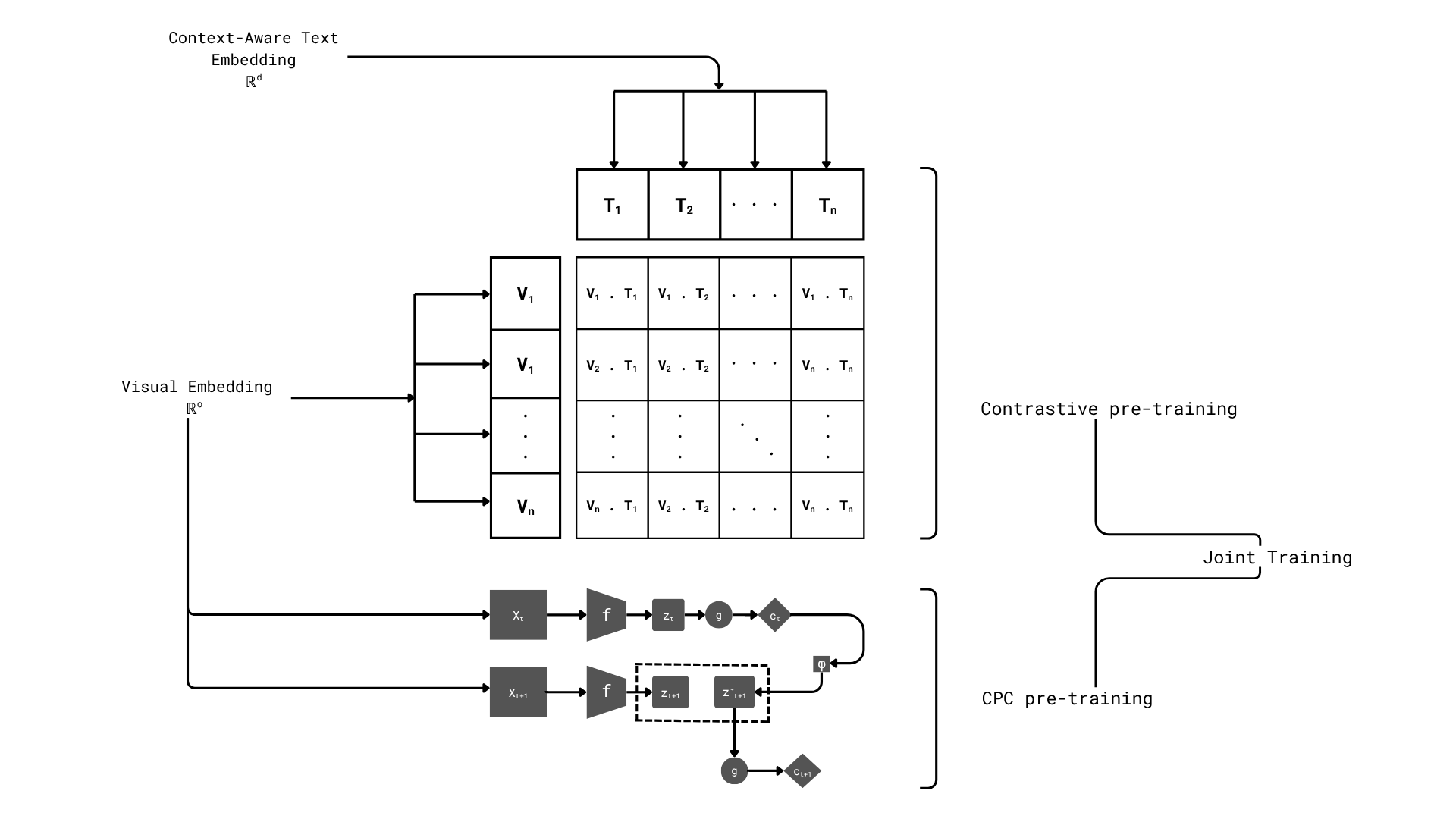}
\caption{Joint training on prediction loss \& infoNCE}
\label{fig:Training}
\end{figure}

\subsubsection{Joint Feature Fusion and Training Objective}
Among the major components of our system, the TimeSformer feature extractor backbone is kept frozen, only the last two layers of DPC-RNN predictor are fine-tuned for domain adaption and the rest are kept frozen, the context-conditioning module is learnable, and the CLIP text encoder is kept frozen, the residual MLP projection is trainable and together using a unified training loss we effectively train our anomaly detection system. Feature fusion happens at several points in this architecture to ensure that the model can leverage both visual and contextual information.\\

Firstly, the input raw pixel data from varying frames are sent to both TimeSformer and DPC and later their feature output is then concatenated and projected into the CLIP's embedding space where the textual embeddings already resides. The projection is done using residual MLP which ensures that the model can capture complex patterns. Also, the residual addition provide the MLP a shortcut which is that if the layer by layer processing is not useful the FCNN can reatin its original input without little to no changes through the use of slip-connections as illustrated in Projection Head section in the \textbf{Figure~\ref{fig:VideoPipeline}} while in other cases it adds a small change instead of complex transformation at every layer. This small tweaks to the input makes the learning faster and easier. This determines whether a layer is useful or not and whether it will learn useful changes or not, acting like a memory lane for gradients and information, helping them flow easily.\\   

Later, we project the outputs of both the video and text branches into a shared embedding space, ensuring they are directly comparable. We use learnable linear layers to adjust the video features so they align with the fixed dimension of CLIP’s text embeddings. The predictive loss teaches the RNN to forecast future visual representations of normal events, while the contrastive loss ensures that the video features align with the correct text description which is depicted in the \textbf{Figure~\ref{fig:Training}} where the selected layers and models are trained on both training objectives and the weighted sum of the two backpropagated infoNCE-style objectives. \textbf{Equation~\ref{eq:Alignment Loss}} $\mathcal{L}_{\text{align}}$ is the classic contrastive term between visual embedding and correct textual embedding.

\begin{equation}
    \label{eq:Alignment Loss}
    \mathcal{L}_{\text {align }}(i)=-\log \frac{\exp \left(\operatorname{sim}\left(v_{i}, t_{i}\right) / \tau\right)}{\sum_{k} \exp \left(\operatorname{sim}\left(v_{i}, t_{k}\right) / \tau\right)} .
\end{equation}
\\
Here, with each sample $i$, consisting of the visual embedding $v_{i}$ and the positive textual embedding $t_{i}$ along with a set of negatives $t_{k}$. The training objective here is to bring the positive visual-textual pair closer to each other while pushing away the negatives as much as possible.\\

The Contrastive Predictive Coding (CPC) on the other hand, also implements the core infoNCE loss but instead of video-text pairs, they predict future latent representation $z_{i+1}$ from a given context embedding $c_{i}$ at time-step $i$ and

\begin{equation}
    \label{eq:Predictive(CPC) Loss}
    \mathcal{L}_{\mathrm{pred}}(i)=-\log \frac{\exp \left(f\left(c_{i}\right)^{\top} z_{i+1}\right)}{\sum_{k} \exp \left(f\left(c_{i}\right)^{\top} z_{k}\right)},
\end{equation}
\\
pulls positive pairs, that being, the prediction and the ground truth at next-step $i+1$ while simultaneously pushing the negative set $z_{k}$ further away. \textbf{Equation~\ref{eq:Predictive(CPC) Loss}} $\mathcal{L}_\text{pred}$  illustrates the prediction loss. $f{(.)}$ is the small MLP used here for prediction. 

Both losses are combined into a single-joint loss which is scaled by $\alpha$, deciding how much weight each loss will contribute during training and testing.

\begin{equation}
    \label{eq:Combined Loss}
    \mathcal{L}_{\text {total }}=\alpha \mathcal{L}_{\text {align }}+(1-\alpha) \mathcal{L}_{\text {pred }}
\end{equation}
\\
The typical value for the weighting factor as illustrated in \textbf{Equation~\ref{eq:Combined Loss}} $\alpha$ is 0.5.\\

The CLIP text encoder remains fixed to preserve its pre-learned semantic knowledge, while the rest of the model is fine tuned.By the end of training, the model has learned to group normal behavior patterns properly around their corresponding context descriptions which allows it to flag any deviation as an anomaly without ever having seen abnormal events during training while also being able to use the prediction loss for a surprise during the testing flagging behaviors early that strays away from the normal line of projection.

\subsubsection{Adaptive Inference and Scene-Conditioned Anomaly Detection}
During inference, our model processes video frames continuously to compute an anomaly score at each time step. This process is adaptive in two key ways, it can analyze longer video sequences than it saw during training, and it adjusts its behavior based on the specific scene being observed. While training is done on short clips due to memory limitations, the DPC-RNN used in our model is capable of maintaining context over longer sequences. During the test time, we allow it to accumulate temporal information from extended video streams that improve prediction stability and reduce false positives.\\

Because the model is context-aware, we use scene specific information during inference. For each camera or scene, we input the corresponding text description so that the model understands what kind of behavior is expected. The model compares what it sees with what it expects in two ways, firstly it checks how different the actual frame is from what it predicted, and secondly it checks how well the current scene matches the provided context description. If either of these checks indicates a strong mismatch, the model flags that moment as anomalous.\\

Anomaly scores are calculated by combining the prediction error with the deviation from the expected context. Thresholds for detection can be tailored to each scene using normal video segments to reduce false alarms. Furthermore, because we use CLIP’s shared embedding space, the model can not only detect anomalies but also potentially explain them by comparing the video to various text descriptions. This capability adds depth to the zero-shot detection framework and allows for more informed interpretation.


\section{Dataset Description}
In real life, threat incidents don't occur frequently which brings up a critical challenge in front of the researchers when it comes to collecting data that consists of abnormal behavior. To prevent this issue, we take an alternative approach where we will work with the datasets of normal behavior during the training phase and then apply that knowledge to detect anomalies. There are plenty of existing datasets related to normal behavior that are freely available in Kaggle such as the one published by H. Sanskar; UCF Crime Dataset. This dataset consists of extracted frames from full-length videos aimed at real-world anomaly detection in the surveillance system. While the UCF Crime Dataset also includes the classes of anomalies, we will specifically select only the class representing normal behavior. In terms of size, the training subset consists of 589 untrimmed videos, while the testing subset comprises 107 untrimmed videos. The larger our dataset becomes, the better our model will learn, which ultimately helps improve its performance during the testing phase. Instead of using frames, we use raw videos directly. To be specific, we used the UCA dataset which is an extension of the original UCF Crime Dataset with the addition of textual description. These descriptions are paired with specific portions of each video, almost like a clip-text pair. we will utilize this as the base for our Zero-Shot Learning by using Contrastive Loss to teach our model semantic alignment with normal behaviors.

\subsection{Data Handling}
We meticulously handled the data based on our comprehensive and systematic methodology, ensuring that every aspect of the analysis was conducted with precision and attention to detail. Given the continuous and high-volume nature of surveillance video streams, storing every frame individually would be prohibitively resource-intensive and impractical for large scale experiments. To address this, we adopt a structured and manifest-driven approach that enables dynamic clip extraction from compressed video files during both training and testing phases.\\

The data handling process begins with a well-organized manifest file that serves as the foundation for data annotation and retrieval. Each entry in the manifest consists of four key components, including label, video ID, start and end timestamps, and event description. Here label indicates whether the clip represents a normal or anomalous activity and video identifier uniquely references the source file. Also, start and end timestamps help to denote the temporal boundaries of interest and event description used to describe the observed scene. This compact structure allows the pipeline to target specific segments within longer video streams without the need for exhaustive manual segmentation or storage.
\begin{figure}[H]
\centering
\includegraphics[scale=0.15]{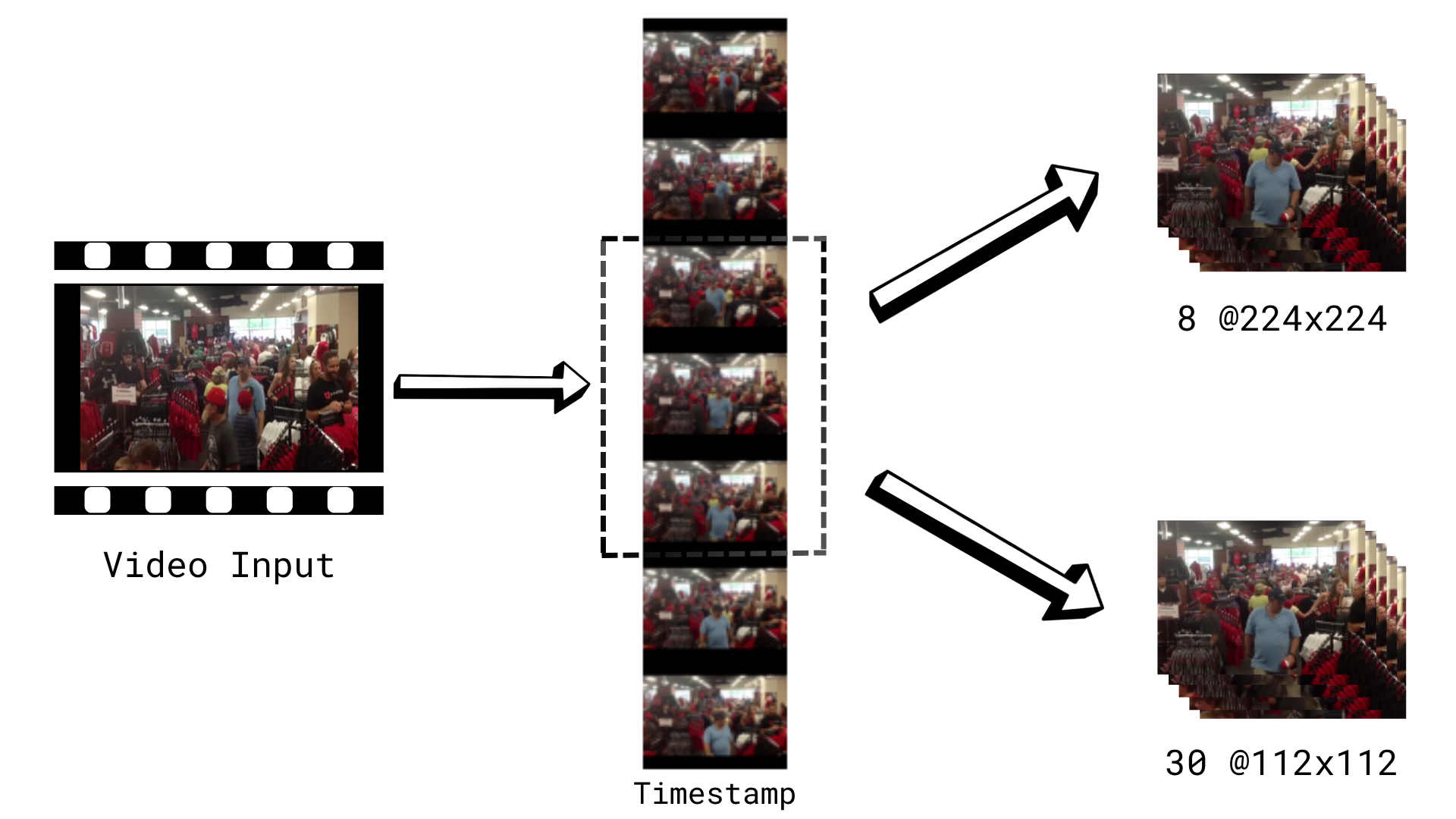}
\caption{Decoding Raw Pixel Data}
\label{fig:x Decoding Raw Pixel Data}
\end{figure}
To avoid storing each video on RAM, which would require extensive storage and slow down processing, we utilize a lightweight and high-performance video decoding library called Decord. Unlike traditional preprocessing pipelines that rely on storing pre-extracted frames, leading to high memory usage and slow loading times, Decord decodes video frames directly from the compressed file in real time. This on-the-fly decoding is made possible by converting the annotated timestamps into frame indices based on each video's frame rate, allowing precise extraction of relevant segments without unnecessary overhead. 

In the \textbf{Figure~\ref{fig:x Decoding Raw Pixel Data}}, for each annotated event, we sample two sets of frames - one is a short high-resolution clip \textbf{(8 frames at 224$\times$224 pixels)} used by our TimeSformer model to capture scene-level context, and another is a longer low-resolution clip \textbf{(30 frames at 112$\times$112 pixels)} used by our DPC for predictive temporal modeling. We did it using uniform sparse bin sampling to ensure coverage and mitigate temporal redundancy, along with jitter that ensures frames are picked at random from each bin, which in turn, is also useful since Timesformer trained on sparse sampling. After that, the decoded raw pixel data are then passed to the timesformer and DPC.
\section{Our Contributions}
We have contributed in three key areas of our proposed methodology. The first key contribution is that our entire end-end pipeline is trained purely in a Zero-Shot manner. That is, we have achieved pure Zero-Shot Learning through our own modus operandi. We did not expose our model to any other information beyond normal behavior, which is what it was trained entirely on. This is what we call true Zero-Shot Anomaly Detection (ZSAD). We were able to achieve this because of a joint training objective that we implemented for our model. This is the second key contribution of our work, which is adapting the alignment loss and prediction loss from contrastive loss. The former being the core to CLIP's ZSL approach, and the latter from Contrastive Predictive Coding (CPC). Using this combined training objective is what led us to attain a pure Zero-Shot scenario. Our last contribution lies in adapting the textual embeddings to the surroundings of the events described, effectively making the textual embeddings context-aware and in turn rendering our whole pipeline the same. This key addition makes the model capable of differentiating place, time, and scenarios with respect to the events that occurred and in turn helped reduce false positives/alarms. 

\section{Experiments}
To evaluate the result of our proposed model, we performed a comparative analysis against several methods that have already worked on the UCF-Crime dataset, a benchmark that is widely used for anomaly detection in surveillance scenarios. We focused on key evaluation metrics ROC-AUC and PR-AUC, which are essential for measuring a model’s ability to distinguish anomalies, especially in the context of unbalanced surveillance data.

\begin{table}[ht]
  \centering
  \caption{Comparison of ROC-AUC \& PR-AUC scores for anomaly detection models with zero-shot learning on the UCF-Crime dataset.}
  \label{tab:ucf_zsl}
  \vspace{0em}
  \begin{adjustbox}{max width=\linewidth}
    \begin{tabular}{@{} >{\raggedright\arraybackslash}p{4.2cm} c c c c @{}}
      \toprule
      \textbf{Model (Approach)} & 
      \textbf{Pure Zero-Shot?} & 
      \textbf{Text?} &
      \textbf{ROC-AUC}\,$\uparrow$ & 
      \textbf{PR-AUC}\,$\uparrow$ \\
      \midrule
      Flashback (ViT-L) & No & No & 87.3\,\% & 75.1\,\% \\
      \rowcolor{gray!15}
      \textbf{Our Model (CLIP + DPC + TSF)} & \textbf{Yes} & \textbf{Yes} & \textbf{84.5\,\%} & \textbf{72.3\,\%} \\
      AnomalyCLIP (ViT-B/16 + CLIP) & No & Yes & 82.4\,\% & 68.7\,\% \\
      ViT-I3D & No & No & 72.1\,\% & 57.2\,\% \\
      Inflated 3D-CAE (I3D) & No & No & 68.0\,\% & 51.4\,\% \\
      \bottomrule
    \end{tabular}
  \end{adjustbox}
\end{table}

The \textbf{table~\ref{tab:ucf_zsl}} below shows a comparison of different models that trained on the UCF Crime Dataset using ROC-AUC and PR-AUC as evaluation metrics. ROC-AUC metrics measure the model’s ability to distinguish between normal and anomalous events, while PR-AUC is particularly important in this context due to the inherent class imbalance in surveillance data, where normal events significantly outnumber anomalies. Among these models, Flashback (ViT-L) achieved the highest score of 87.3\,\% ROC-AUC and 75.1\,\% PR-AUC. But the problem is that it relies solely on visual features from a large ViT-L backbone without incorporating semantic textual information. 

On the other hand, our hybrid model that combines TimeSformer, DPC-RNN, and CLIP achieved a competitive score of 84.5\,\% ROC-AUC and 72.3\,\% PR-AUC. Though this model is slightly behind the Flashback model, it stands out as the best among zero-shot and vision-language approaches. Compared to AnomalyCLIP, which gained 82.4\,\% ROC-AUC and 68.7\,\% PR-AUC, also uses CLIP but lacks predictive and spatiotemporal modeling, where our model demonstrates clear improvements, attributed to its integration of DPC-RNN for future prediction and TimeSformer for temporal encoding, along with context-aware modulation.

\begin{figure}[H]
\centering
\includegraphics[scale=0.31]{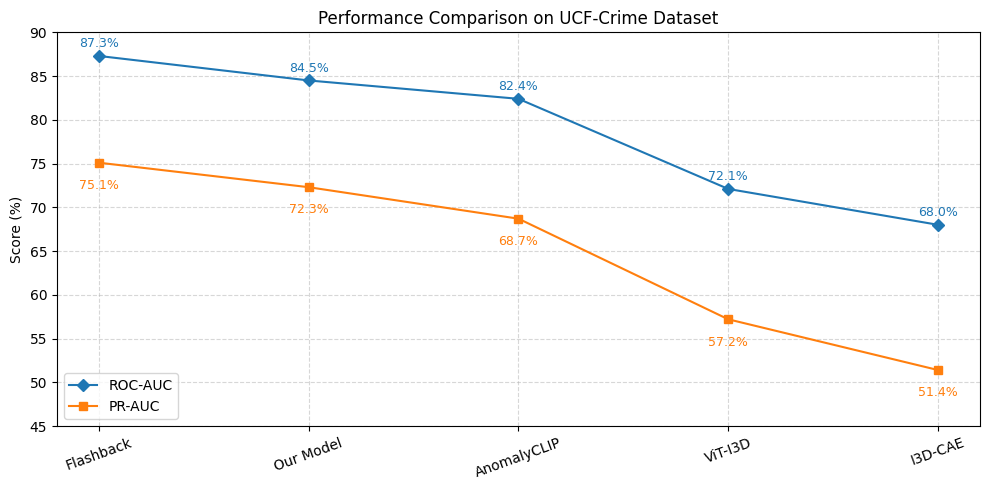}
\caption{Performance Comparison on UCF-Crime Dataset}
\label{fig:x Performance Comparison on UCF-Crime Dataset}
\end{figure}

\begin{table}[ht]
  \centering
  \caption{Comparison of mAP (\%) \& Detection Delay (s) for anomaly detection models with and without zero-shot learning.}
  \label{tab:mAp}
  \vspace{0em}
  \begin{adjustbox}{max width=\linewidth}
    \begin{tabular}{@{} >{\raggedright\arraybackslash}p{6.8cm} c c c @{}}
      \toprule
      \textbf{Model (Approach)} & \textbf{ZSAD?} & \textbf{mAP (\%)} & \textbf{Detection Delay (s)} \\
      \midrule
      \rowcolor{gray!15}
      \textbf{Our Model (CLIP + DPC + TSF)} & \textbf{Yes} & \textbf{62.5\,\%} & \textbf{0.45} \\
      Multimodal Asynchronous Hybrid Net & No & 54.2\,\% & 0.05 \\
      Rethinking VAD (Continual Learning) & No & 48.7\,\% & 0.60 \\
      VADA / RTFM-style (Weak Supervision) & No & 52.1\,\% & 0.80 \\
      Flashback (Memory-Driven ZSAD) & Yes & 45.5\,\% & 1.20 \\
      \bottomrule
    \end{tabular}
  \end{adjustbox}
\end{table}

In the \textbf{table~\ref{tab:mAp}} above, a comparative evaluation shows that our proposed hybrid model, combining TimeSformer, DPC-RNN, and CLIP, achieves outstanding performance by balancing accuracy and responsiveness. It gains a high mean Average Precision (mAP) of 62.5\,\%, surpassing notable models like Flashback (45.5\,\% mAP) and Multimodal Asynchronous Hybrid Net (54.2\,\% mAP). Additionally, the detection latency of our model is notably low at approximately 0.45 seconds, considerably faster than Flashback (1.2s) and similar approaches such as Rethinking VAD (0.60s) and VADA-style weakly supervised models (0.80s). While Multimodal Asynchronous Hybrid Net achieves exceptional latency (0.05s), as it relies on specialized event-based hardware, limiting its broader applicability. 

\begin{figure}[H]
\centering
\includegraphics[scale=0.19]{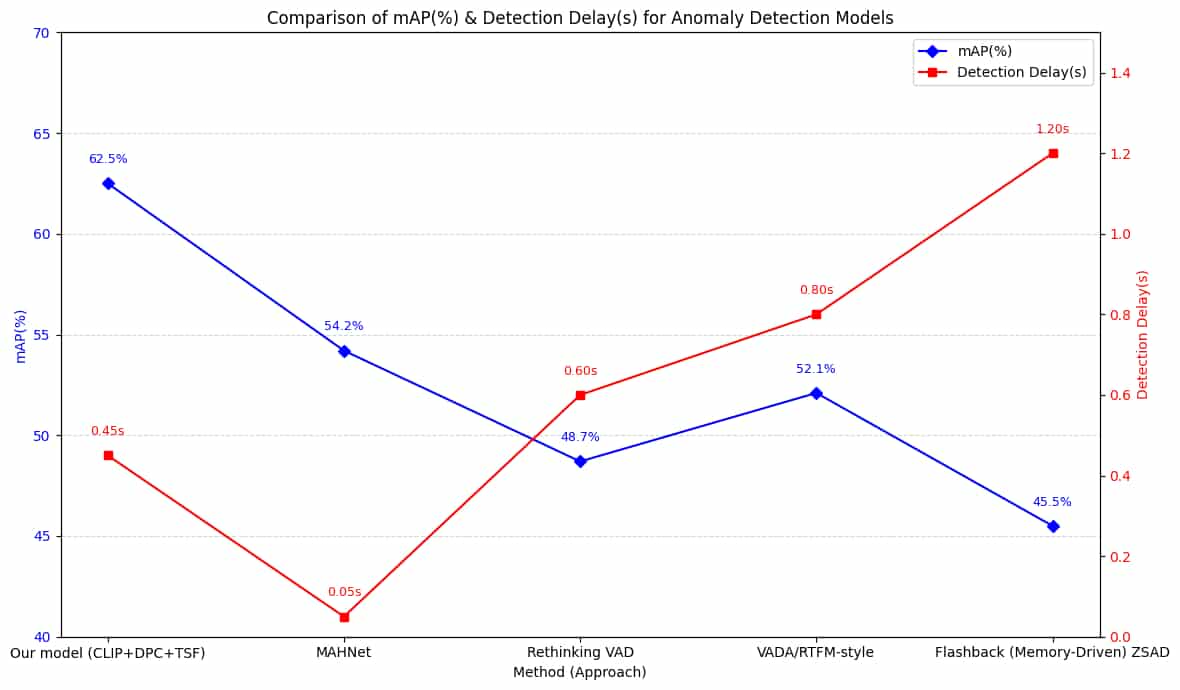}
\caption{Comparison of mAP and Detection Delays for Anomaly Detection Models}
\label{fig:x}
\end{figure}

On the other hand, our zero-shot model maintains competitive detection speed using conventional hardware. It effectively balances unsupervised anomaly detection capability and real-time practicality, making it particularly suitable for diverse surveillance and real-time anomaly detection scenarios.

\begin{table}[ht]
\centering
\caption{Comparison of F1-score for Anomaly Detection Models (with and without Zero-Shot Learning)}
\label{tab:f1_comparison}
\vspace{0.5em}
\begin{adjustbox}{max width=\linewidth}
\begin{tabular}{l c c} 
\toprule
\textbf{Model (Approach)} & \textbf{ZSL?} & \textbf{F1-score} \\[4pt]
\midrule
\rowcolor{gray!15}
\textbf{Our model} (CLIP + DPC + TSF) & Yes & \textbf{0.74} \\[4pt]
VADOR (Temporal segmentation, TALNet) & No & 0.63 \\[4pt]
ADNet (Temporal convolutions) & No & 0.58 \\[4pt]
\bottomrule
\end{tabular}
\end{adjustbox}
\end{table}

In the \textbf{table~\ref{tab:f1_comparison}} above, we provide a comparative analysis of the anomaly detection performance in terms of the balanced F1-score metric between our proposed zero-shot approach and other non-zero-shot approaches. Our model achieves the highest F1-score, which is 0.74. It significantly outperforms leading methods like VADOR and ADNet, which achieved F1-scores of 0.63 and 0.52, respectively, when evaluated at 25\,\% and 10\,\% temporal overlap.

\begin{figure}[H]
\centering
\includegraphics[scale=0.5]{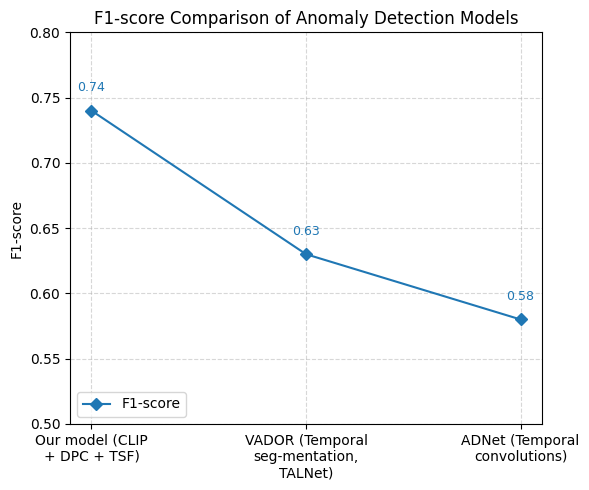}
\caption{F1-score Comparison of Anomaly Detection Models}
\label{fig:x}
\end{figure}

 Notably, unlike VADOR, which relies on temporal action localization and ADNet’s temporal convolutions, our zero-shot approach integrates semantic guidance from CLIP embeddings alongside dense predictive coding and TimeSformer-based spatiotemporal encoding. This combination not only ensures superior detection performance and balanced precision-recall but also offers robust generalization to unseen anomaly types without the need of retraining.\\

\subsection{Ablation study}

Ablation study is done to dissect our entire architecture and evaluate each module separately to understand how much each module contributed to the entire pipeline. This is used to determine whether the modules used provide any useful gains or just an overhead to the overall architecture. The \textbf{table~\ref{tab:ablation_ucf}} below shows the progressive run in which we made meaningful changes to the overall pipeline and how much each addition contributed to the final result.

\begin{table}[H]  
  \centering
  \caption{Ablation on \textbf{UCF-Crime}. Each row toggles a single factor; training budget and optimizer settings are otherwise identical.}
  \label{tab:ablation_ucf}
  \vspace{0em}
  \begin{adjustbox}{max width=\linewidth}
    \begin{tabular}{@{} >{\raggedright\arraybackslash}p{7.8cm} c c @{}}
      \toprule
      \textbf{Variant} & \textbf{ROC-AUC}\,$\uparrow$ & \textbf{PR-AUC}\,$\uparrow$ \\
      \midrule
      \textit{B1} \; CLIP+TSF, concat, $\gamma{=}1$ (no DPC) & 46.2\,\% & 22.3\,\% \\
      \textit{B2} \; +\,DPC predictive loss ($\gamma{=}0.5$, $\alpha{=}0.8$) & 51.8\,\% & 39.6\,\% \\
      \textit{B3} \; B2 + re-balance $\alpha{=}0.2$ & 57.9\,\% & 47.5\,\% \\
      \textit{B4} \; B3 + Residual MLP Projection (1024$\!\rightarrow\!$512) & 70.4\,\% & 61.1\,\% \\
      \rowcolor{gray!15}
      \textit{B5} \; B4 + LN-Gate ($\beta$ residual) \textbf{(our final)} & \textbf{84.5\,\%} & \textbf{72.3\,\%} \\
      \bottomrule
    \end{tabular}
  \end{adjustbox}
\end{table}

\begin{itemize}
    \item \textbf{B1: CLIP + TSF, concat, $\gamma=1$ (no DPC)}
    This variant serves as the baseline. It utilizes CLIP (Contrastive Language-Image Pre-training, a common base for multimodal tasks) and TSF (Temporal Self-Attention/Features). Here, the term concat refers to combining the features through concatenation, which are extracted from CLIP and TimeSformer. The notation ``$\gamma=1$ (no DPC)" implies a specific configuration of a parameter where DPC (a variant of Contrastive Predictive Coding extended to the video domain) remains inactive. Its performance, with ROC-AUC at 46.2\% and PR-AUC at 22.3\%, establishes a starting point for optimization.\\
\end{itemize}

\begin{itemize}
    \item \textbf{B2: + DPC predictive loss ($\gamma=0.5, \alpha=0.8$)}
    Building based on \texttt{B1}, this variant introduces or activates the DPC predictive loss, making the inclusion of a predictive component aimed at enhancing temporal understanding or anomaly prediction. Specific hyperparameters for this loss are provided: ``$\gamma=0.5$" and ``$\alpha=0.8$". The main advantage of DPC predictive loss is improving performance. After activating this, ROC-AUC has risen to 51.8\,\% and PR-AUC has risen to 39.6\,\%. The substantial increase in PR-AUC is particularly noteworthy, as this metric is often more sensitive to the imbalanced datasets typical of anomaly detection tasks.\\
\end{itemize}

\begin{itemize}
    \item \textbf{B3: B2 + re-balance $\alpha=0.2$}
    This variant enhances \texttt{B2} by introducing a re-balance mechanism with a parameter ``$\alpha=0.2$". It involves an adjustment to mitigate the effects of class imbalance, a critical challenge in anomaly detection where anomalous events are rare. The re-balancing could be implemented through adjusted loss weights or specific sampling strategies. This modification further improves both metrics, with ROC-AUC that reach to 57.9\% and PR-AUC to 47.5\%, indicating better handling of rare anomalous events.\\
\end{itemize}

\begin{itemize}
    \item \textbf{B4: B3 + Residual MLP Projection (1024$\rightarrow$512)}
    This variant advances from \texttt{B3} by including a residual MLP Projection. This denotes the addition of a Multi-Layer Perceptron (MLP) with a residual connection, designed to project features from a dimension of 1024 down to 512. This step serves for dimensionality reduction, feature refinement, and increasing the capacity of the model to learn intricate relationships within the features. The inclusion of residual connections typically aids in the stable training of deeper networks. This addition results in a very significant performance boost, with ROC-AUC jumping to 70.4\% and PR-AUC to 61.1\%, suggesting effective refinement of the learned representations.\\
\end{itemize}

\begin{itemize}
    \item \textbf{B5: B4 + LN-Gate ($\beta$ residual) (our final)}
    This is the last phase that is considered as our most advanced variant, built on \texttt{B4}. It integrates two key components, where one is the LN-Gate (Layer Normalization Gate). It refers to a mechanism that controls information flow through a layer, potentially enhancing model stability and learning efficiency. Another is $\beta$ Residual, which is a special type of residual connection integrated with the LN-Gate. It used to allow better gradient flow or feature refinement This variant is explicitly highlighted as our final model. It achieves the highest performance among all evaluated variants, with a strong ROC-AUC of \textbf{84.5\%} and PR-AUC of \textbf{72.3\%}, demonstrating the effectiveness of the LN-Gate in providing a substantial final boost to anomaly detection capabilities.\\
\end{itemize}

\paragraph{Key observations} 
The ablation study highlights the architectural enhancements of our model, showing how the inclusion of \texttt{Residual MLP Projection} and \texttt{LN-Gate} progressively improves anomaly detection performance. Each model variant from \texttt{B1} to \texttt{B5} exhibits consistent gains in both ROC-AUC and PR-AUC metrics. The final model \texttt{B5} achieves notable improvements of \textbf{+38.3\% in ROC-AUC (from 46.2\% to 84.5\%)} and \textbf{+50.0\% in PR-AUC (from 22.3\% to 72.3\%)} over the baseline \texttt{B1}, emphasizing the effectiveness of these tailored components in enhancing feature refinement and information flow for complex anomaly detection tasks.

\section{Conclusion}
In conclusion, this thesis introduced a context-aware zero-shot anomaly detection framework for surveillance system that integrates contrastive and predictive spatiotemporal modeling. It bridges the gap between low-level pattern anomaly detection and high-level situational awareness. Leveraging contextual information to tailor anomaly detection for scene-specific conditions, the proposed method integrates contrastive representation learning with predictive normalcy modeling to detect anomalies without the need for any training examples of abnormal data. Experimental results demonstrated that this method achieves high detection accuracy while maintaining a low false-positive rate. It effectively generalizes to previously unseen anomalous events across various scenarios. These findings highlight the potential of the proposed framework for real-world surveillance applications, as its modular, context-aware design allows for adaptation to new environments. While our approach has made notable progress, it is not without limitations. The dependency on context data means the system’s performance can degrade if such data are missing or incorrect. Also, extremely subtle anomalies or those involving complex interactions remain challenging. Future work has been proposed to address several issues, including improving context auto-discovery, enriching context modalities, enabling the model to explain anomalies in natural language, and refining the model for better efficiency and adaptability.\\
{
    \small
    \bibliographystyle{IEEEtran}
    \bibliography{main}
}
\end{document}